\documentclass{article}
\usepackage{iclr2027_conference,times}

\usepackage{amsmath,amsfonts,bm}

\def\eqref#1{equation~\ref{#1}}

\def\1{\bm{1}}

\DeclareMathAlphabet{\mathsfit}{\encodingdefault}{\sfdefault}{m}{sl}
\SetMathAlphabet{\mathsfit}{bold}{\encodingdefault}{\sfdefault}{bx}{n}

\usepackage{amsmath,amssymb}
\usepackage{booktabs}
\usepackage{array}
\usepackage{graphicx}
\usepackage{microtype}
\usepackage{xcolor}
\usepackage{flafter}
\usepackage{float}
\usepackage{wrapfig}
\usepackage{hyperref}
\usepackage{url}
\hypersetup{hidelinks}
\usepackage{multirow}
\usepackage{booktabs}
\usepackage{graphicx}
\usepackage{multirow}
\usepackage{amssymb}
\usepackage{pifont}

\newcommand{\xmark}{\ding{55}}
\newcommand{\method}{\textsc{HarnessSQL}}

\newif\ifshownotes
\shownotesfalse
\definecolor{noteJP}{RGB}{180,110,20}
\ifshownotes
  \newcommand{\jipeng}[1]{\textcolor{noteJP}{[JP: #1]}}
\else
  \newcommand{\jipeng}[1]{}
\fi

\title{\method: Harness-Native Training for SQL Agents in Realistic Database Environments}

\author{%
{\bfseries Haolin Yang$^1$\thanks{Equal contribution.} \quad
Jipeng Zhang$^2$\footnotemark[1] \quad
Zhiqiang Qian$^1$ \quad
Jian Xie$^3$ \quad
Shuaishuai Gong$^4$} \\
{\bfseries Yike Guo$^1$\thanks{Corresponding authors.} \quad
Sirui Han$^1$\footnotemark[2]} \\
{\normalfont\small
$^1$Hong Kong University of Science and Technology \quad
$^2$Microsoft Research} \\
{\normalfont\small
$^3$Tsinghua University \quad
$^4$University of Macau}
}

\iclrfinalcopy
\begin{document}

\maketitle
\lhead{Preprint}

\begin{abstract}
Text-to-SQL models are commonly trained to map questions directly to static queries, whereas real-world database agents operate through stateful, multi-turn interaction with live databases—inspecting schemas, executing probe queries, diagnosing errors, and revising hypotheses. This creates a critical train--deploy mismatch, as the execution harness that mediates this interaction is introduced only at inference time. To bridge this gap, we propose \method{}, a harness-native post-training framework that preserves the full interaction structure throughout both supervised fine-tuning and reinforcement learning. \method{} builds isolated, executable database environments paired with hidden execution oracles, rolls out teachers directly inside the target SQL harness, and retains only verified trajectories for full-sequence SFT, followed by execution-reward RL. Across Spider~2.0-SQLite, \method{} dramatically boosts the execution accuracy of compact models, raising Qwen3-8B from 15.5\% to 45.2\% and Qwen3-14B from 22.2\% to 54.8\%, while transferring effectively to out-of-distribution interactive benchmarks such as BIRD-Interact and LiveSQLBench. Our findings demonstrate that training database agents directly within their execution harness is essential for mastering complex, long-horizon database workflows. 
Code is released at \url{https://github.com/YangHaolin0526/HarnessSQL}
\end{abstract}

\section{Introduction}
\begin{figure*}[ht]
\centering
\vspace{-0.2cm}
\includegraphics[width=\textwidth]{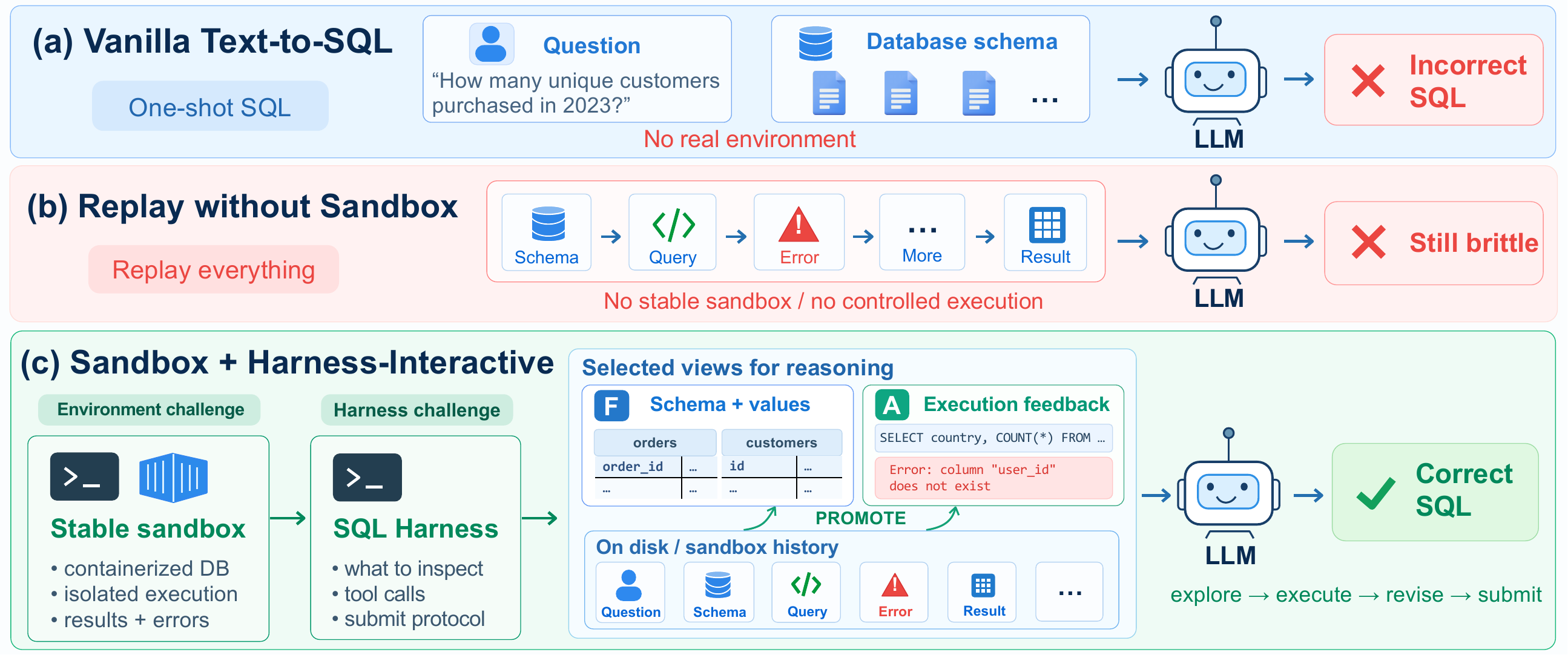}
\vspace{-0.7cm}
\caption{From one-shot Text-to-SQL to harness-interactive database agents. A stable sandbox and structured SQL harness enable iterative exploration, 
execution feedback, revision, and submission.}
\label{fig:traditional-workflow}
\end{figure*}
Text-to-SQL has traditionally been formulated as a direct mapping from a natural-language question and database schema to an executable SQL query~\citep{yu2018spider,li2023bird}. This abstraction has enabled substantial progress, but it increasingly diverges from how modern systems solve realistic database tasks. Benchmarks such as Spider2.0\citep{fang2025spider2} place agents in large, unfamiliar database environments where the information required to solve a task cannot always be captured in a static prompt. An agent may need to inspect tables, examine schemas and values, execute intermediate queries, diagnose errors, and revise its solution over multiple turns. Recent systems have consequently begun to treat Text-to-SQL as an interactive problem, mirroring the broader rise of autonomous data agents~\citep{zhu2026insighteval} that use tools and execution feedback to reason over databases rather than generating a query in a single pass~\citep{deng2025reforce,li_swe-sql_2025}. This shift raises a corresponding question: \emph{how should such SQL agents be trained?}

Figure~\ref{fig:traditional-workflow} illustrates why interactive SQL agents require more than simply adding tools at inference time. Even for a seemingly simple request such as ``How many unique customers purchased in 2023?", the agent may not know in advance which tables, keys, or values are relevant\citep{pham2026flexsqlflexibleexplorationexecution}. It may first inspect the schema, issue a plausible query, encounter an execution error such as referencing a nonexistent \texttt{user\_id} column, and then use the returned database state to revise its hypothesis before submitting a correct query. This process exposes three coupled challenges. First, the environment must be executable and reproducible: the agent needs access to authentic schemas, values, query results, and runtime errors rather than a static textual approximation. Second, the harness itself shapes the policy: which inspection and execution tools are available, what observations are exposed, and how submission and termination are defined determine the sequence of decisions the model can make. Third, the supervision must capture the interaction process rather than only the terminal SQL: a correct final query does not teach the model when to inspect the schema, how to interpret an execution failure, or how to recover from an incorrect intermediate hypothesis\citep{guo-etal-2026-mtsql,shi2025legalreasoner}. Recent interactive SQL benchmarks and harness studies increasingly expose these requirements at evaluation time\citep{xu2025mtirsqlmultiturntoolintegratedreasoning,yao2026harnessbench}, yet most SQL post-training still focuses on static question–SQL supervision or terminal execution correctness. This creates a fundamental train–deploy mismatch: models are trained to produce answers, but deployed as policies that must explore, execute, revise, and terminate through a concrete database harness.

To resolve these challenges, we introduce \method{}, a harness-native post-training framework that aligns the supervision of database agents directly with their runtime execution interface. \method{} consists of three core components: 
\textbf{1. Executable SQL Environment Construction.} We instantiate isolated database sandboxes that provide authentic schemas, query execution, runtime feedback, and hidden executable verification, enabling agent trajectories to be generated and evaluated against real database states.
\textbf{2. SQL-Specific Harness Adaptation.} Rather than directly adopting a generic agent harness, we adapt its interaction interface and execution protocol to the characteristics of SQL tasks. The resulting harness exposes task-relevant observations and actions while reducing unnecessary interaction complexity, making teacher rollouts more reliable and the resulting trajectories easier for compact models to learn from.
\textbf{3. Harness-Native Post-Training.} We collect oracle-verified full interaction trajectories directly inside the adapted harness and use them for supervised fine-tuning, followed by execution-reward reinforcement learning in the same environment. This two-stage training process teaches the model not only to produce correct SQL, but also to acquire the complete interaction policy for exploration, execution, recovery, and submission.

Empirical results demonstrate that harness-native training substantially elevates compact models. On Spider~2.0-SQLite, \method{} improves Qwen3-8B execution accuracy from 15.5\% to 45.2\% and Qwen3-14B from 22.2\% to 54.8\% after SFT and RL. Furthermore, the learned policies transfer effectively to out-of-distribution interactive environments, reaching 9.0\%/11.3\% on BIRD-Interact Mini and 24.8\%/28.5\% on LiveSQLBench Base-Lite for 8B/14B models. Controlled comparisons confirm that harness-native post-training significantly outperforms conventional non-harness fine-tuning, highlighting the importance of aligning training with the deployment interaction interface.

Our contributions are threefold: 
\textbf{(1) Problem Formulation:} We identify a train–deploy mismatch in database agents and formulate interactive Text-to-SQL as long-horizon policy learning grounded in an executable harness, highlighting the coupled roles of environment, interaction interface, and trajectory-level supervision. 
\textbf{(2) \method{} Framework:} We develop a harness-native pipeline spanning sandbox environment construction, dedicated SQL harness design, and end-to-end full-trajectory post-training via SFT and execution-reward RL.
\textbf{(3) State-of-the-Art Compact Agents:} We show that \method{}substantially improves 8B/14B open-weight models on Spider~2.0 and transfers to out-of-distribution interactive benchmarks, demonstrating the effectiveness of training database agents natively within their deployment environment.

\section{Problem Setting}
\vspace{-3pt}

\subsection{SQL agents as partially observed policies}
We formalize interactive Text-to-SQL as a partially observed Markov decision process (POMDP) mediated by an execution harness. A task instance is defined as a tuple $x = (q, \mathcal{D}, h, v)$, where $q$ is the natural-language user instruction, $\mathcal{D}$ denotes the underlying database instance, $h$ specifies the execution harness, and $v$ is a hidden executable oracle.

At each turn $t \in [1, T]$, the agent receives an observation $o_t \in \mathcal{O}$ emitted by the harness and generates an action $a_t \in \mathcal{A}$ according to its policy $\pi_\theta(a_t \mid s_t)$, where the visible state $s_t = (q, a_1, o_1, \dots, a_{t-1}, o_{t-1})$ comprises the dialogue history. The action $a_t$ consists of natural-language reasoning accompanied by either a tool call (e.g., schema lookup or query execution) or a terminal submission. An entire interaction episode is represented as:
\begin{equation}
    \tau = (q, a_1, o_1, a_2, o_2, \dots, a_T, o_T, y),
\end{equation}
where $y$ is the final submitted SQL artifact extracted from $a_T$. Crucially, the agent never observes the full ground-truth state of $\mathcal{D}$ directly; instead, $o_t = h(\mathcal{D}, a_t)$ is a harness-dependent projection bounded by system protocols (such as row truncations, query timeouts, and error message formatting).

Task success is evaluated over the complete trajectory by a joint reward function:
\begin{equation}
    R(\tau; x) = R_{\mathrm{result}}(y, v) \cdot R_{\mathrm{protocol}}(\tau, h).
\end{equation}
Here, $R_{\mathrm{result}}(y, v) \in \{0, 1\}$ evaluates the denotational execution equivalence of the submitted artifact $y$ against the hidden verifier $v$ over $\mathcal{D}$, rather than relying on surface-form string matching. $R_{\mathrm{protocol}}(\tau, h) \in \{0, 1\}$ verifies compliance with harness invariants, including read-only query constraints, adherence to action schemas, absence of repeated redundant queries, and termination within the step budget $T$.
\begin{figure*}[t]
\centering
\includegraphics[width=\textwidth]{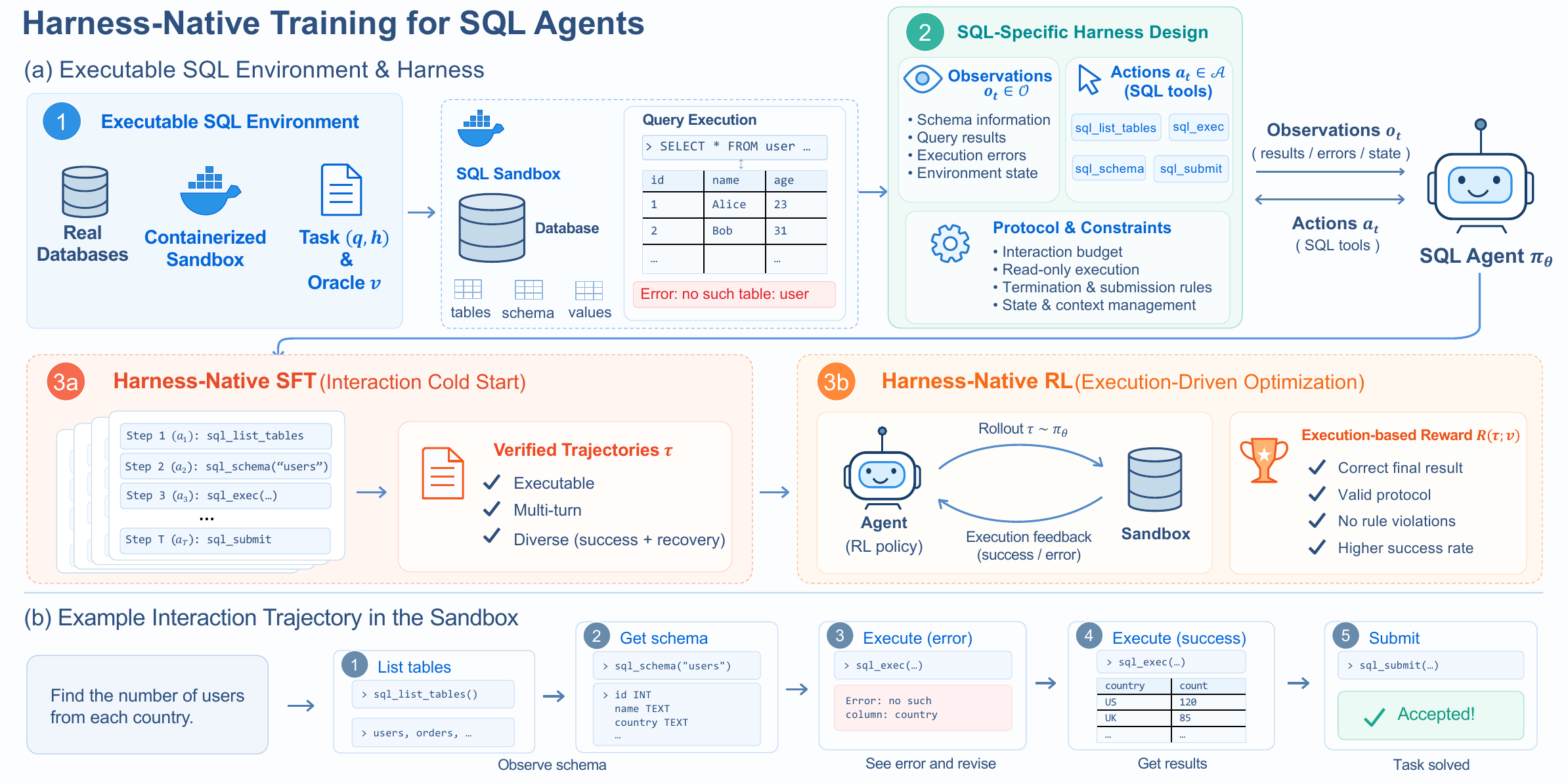}
\caption{
Overview of \method{}, our harness-native post-training framework for interactive SQL agents.
(a) Databases and hidden execution oracles are encapsulated into isolated, containerized sandboxes, where a SQL-specific harness mediates agent--environment interaction through four actions: table discovery (\texttt{sql\_list\_tables}), schema inspection (\texttt{sql\_schema}), query execution (\texttt{sql\_exec}), and submission (\texttt{sql\_submit}). The harness exposes schemas, query results, and execution errors while enforcing interaction and execution constraints. Harness-native SFT provides an interaction cold start from verified multi-turn trajectories, after which harness-native RL further optimizes the policy through online sandbox interaction and execution-based rewards.
(b) An example trajectory illustrates the resulting workflow: the agent discovers tables, inspects the schema, executes a query, observes an execution error, revises its solution, and submits the verified result.
}
\label{fig:pipeline}
\end{figure*}
\subsection{The Train--Deploy Mismatch}
Existing SQL post-training frameworks predominantly conceptualize Text-to-SQL as a one-shot conditional generation task. Formally, conventional supervision maximizes the likelihood of the golden query $y^*$ conditioned on the prompt and a static schema description $\mathcal{S}$:
\begin{equation}
    \mathcal{L}_{\mathrm{static}}(\theta) = - \log \pi_\theta(y^* \mid q, \mathcal{S}).
\end{equation}
Even recent reasoning-oriented methods that incorporate chain-of-thought traces $c$ still formulate the objective as $\pi_\theta(c, y^* \mid q, \mathcal{S})$, terminating statically without environment interaction. 

When deploying such a model as an interactive agent, a severe distributional and structural mismatch emerges across three orthogonal dimensions:

\textbf{1. Marginalizing out intermediate states.} By treating the final query $y$ as the sole optimization target, static training marginalizes over the interaction trajectory:
\begin{equation}
p(y \mid q,S)
=
\sum_{\tau \setminus \{y\}}
p(\tau \mid q,S).
\end{equation}
Yet in interactive environments, transitions $p(o_t \mid s_t,a_t)$ are determined by live database execution.
Discarding these intermediate states removes the procedural policy for exploration, execution, and termination, corroborating recent findings that intermediate trajectory traces, rather than
  terminal targets alone, causally govern model generalization~\citep{wen2026notjust}.

\textbf{2. Policy blindness to execution feedback.}
A deployed SQL agent operates under closed-loop control:
\begin{equation}
a_t \sim \pi_\theta(\cdot \mid q,a_1,o_1,\ldots,a_{t-1},o_{t-1}).
\end{equation}
Static supervision never exposes the model to runtime observations such as execution errors or empty results, creating exposure bias when such feedback must be handled at inference time.

\textbf{3. Harness-oblivious action spaces.}
Bridging this mismatch requires \emph{harness-native training}: explicitly optimizing $\pi_\theta(a_t \mid s_t)$ over trajectories $\tau$ collected directly within the deployment harness $h$, driven by execution-grounded rewards $R(\tau; x)$.
\vspace{-4pt}
\section{Methodology}
Figure~\ref{fig:pipeline} illustrates the \method{} pipeline. To resolve the environment, harness, and supervision mismatches identified in Section~2, \method{} couples hermetic sandbox construction with an SQL-specific interaction contract, synthesizing verified trajectories for full-sequence SFT and execution-reward RL.
\subsection{Executable SQL Environment Construction}
\textbf{Catalog grounding and task synthesis.}
For each target database, we extract an execution-grounded catalog directly from the live engine, capturing runtime schemas, sample column values, and valid join paths. Rather than pairing questions with post-hoc queries, task construction follows an \emph{SQL-first} blueprinting approach: the generator samples an executable SQL query adhering to difficulty controls (Appendix~\ref{app:difficulty-control}) and derives both the natural-language prompt and the execution oracle from the ground-truth relation.

\textbf{Hermetic sandboxing and deterministic gating.}
Each generated task packages public instruction fields $(q, h)$ and a hidden oracle verifier $v$. Tasks are instantiated inside isolated container sandboxes enforcing strict read-only execution, single-statement dialect constraints, and child-process deadlines (60s). Candidate blueprints pass through automated deterministic gates that prune syntax errors, empty relations, degenerate joins, and uninformative edge cases. Mutation tests further perturb queries to ensure semantic discriminability before rollout.
\vspace{-4pt}
\subsection{SQL-Specific Harness Adaptation}
Rather than exposing an unbounded coding environment, we specialize the DSH substrate~\citep{deepseekharness2026} to define a minimal, database-aligned state--action contract:

\textbf{Action space ($\mathcal{A}$):} The model-visible tool interface is restricted to four primitives: \texttt{sql\_list\_tables} (table and row enumeration), \texttt{sql\_schema} (DDL inspection and sample values), \texttt{sql\_exec} (bounded read-only query trial), and \texttt{sql\_submit} (terminal artifact submission). Unrelated shell, filesystem, and code-runtime tools are disabled.
\textbf{Observation and protocol middleware ($\mathcal{O}$):} The environment plane injects runtime guards around every turn, including deterministic row truncation to bound observation length, repeated-call reminders to mitigate loop deadlocks, and pressure-triggered context compaction for long-horizon episodes to prevent distraction and memory degradation under noisy multi-turn observations~\citep{liang2026lost,zhang2026memcore}. Detailed lifecycle hooks are elaborated in Appendix~\ref{app:harness}.
\vspace{-3pt}
\subsection{Harness-Native Post-Training}
\vspace{-3pt}
\textbf{Supervised fine-tuning (cold start).}
We collect demonstration experience by deploying an expert teacher directly within the target harness. For each task, the teacher interacts with the live database solely through the four-tool contract without access to the hidden oracle. Trajectories that pass denotational verification and protocol checks are preserved, yielding 2,512 verified episodes across 79 databases. We train the student policy via full-trajectory SFT:
\begin{equation}
    \mathcal{L}_{\mathrm{SFT}}(\theta) = - \sum_{t=1}^T \log \pi_\theta(a_t \mid s_t), \quad \text{where } s_t = (q, a_1, o_1, \dots, a_{t-1}, o_{t-1}).
\end{equation}
Crucially, tokens originating from the harness observations $o_t$ are masked out of the loss, training the model to orchestrate reasoning, tool invocation, and error recovery conditioned on runtime feedback.

\textbf{Reinforcement learning (execution optimization).}
To encourage the agent to discover autonomous exploration and self-correction paths beyond teacher demonstrations, we continue training via reinforcement learning in the identical harness environment. We optimize the policy using Decoupled Clip and Dynamic Sampling Policy Optimization (DAPO)~\citep{yu2025dapo}. For each task rollout, the hidden verifier evaluates the submitted artifact against the oracle, providing a sparse, execution-based terminal reward $R(\tau; x) \in \{0, 1\}$, in contrast to dense
  step-level process reward models~\citep{chan2025boosting}. Token-level policy gradients are applied across all generated assistant segments. By preserving identical tool semantics, observation boundaries, and termination protocols across both SFT and RL, the policy natively masters the closed-loop decision process required at deployment. Complete training objectives and hyperparameters are detailed in Appendix~\ref{app:training}.
\vspace{-3pt}
\section{Experimental Setup}
\vspace{-3pt}
\textbf{Corpus and Data Synthesis.}
We construct our training corpora from 79 databases originating from Spider~2.0-Lite (30 SQLite) and Spider~2.0-DBT (49 DuckDB ported to SQLite with full schema and column parity). For SFT, collecting verified teacher rollouts yields 2,512 multi-turn trajectories. For reinforcement learning, our training pool comprises 2,700 synthesized question--SQL prompts and 100 validation instances. Detailed database mappings, difficulty stratification, and corpus statistics are provided in Appendix~\ref{app:sft-corpus}.

\textbf{Harness Deployment.}
Evaluations and rollouts strictly follow the 4-tool \textsc{DSH-SQL} contract defined in Section~3.2. Each task runs in a fresh, isolated child process with read-only database bindings, a 60-second execution deadline, and bounded observation buffers. Tasks terminating without an explicit \texttt{sql\_submit} call are marked as failures. Full runtime architecture, middleware hooks, and activation logs are detailed in Appendix~\ref{app:harness}.

\textbf{Model Training.}
We evaluate \method{} across two compact base models: Qwen3-8B and Qwen3-14B. Models are trained sequentially via harness-native full-trajectory SFT and execution-reward RL using DAPO. Optimization is conducted over isolated DSH environments. Specific training hyperparameters, sequence length caps, and compute configurations are deferred to Appendix~\ref{app:training}.

\textbf{Evaluation Benchmarks.}
We evaluate execution accuracy under both in-domain and out-of-distribution (OOD) settings: (i) \textbf{Spider~2.0-SQLite} (135 complex tasks from Spider~2.0-Lite) as the in-domain benchmark; (ii) \textbf{BIRD-Interact Mini}~\citep{huo2025bird-interact} (under the A-Interact interactive protocol with a GPT-4o user simulator) for multi-turn user/environment interaction transfer; and (iii) \textbf{LiveSQLBench Base-Lite SQLite}~\citep{livesqlbench2026} for cross-database and complex reasoning generalization. Comprehensive benchmark specifications are given in Appendix~\ref{app:benchmarks}.

\textbf{Baseline Models and Evaluation Protocols.}
We compare against three groups of baselines in Table~\ref{tab:main}. \textbf{General-purpose models} include Qwen2.5-Coder-32B, providing coding-oriented and reasoning-capable references. \textbf{SQL-specialized models} include OmniSQL\citep{li2025omnisql} and Arctic-Text2SQL\citep{yao-etal-2026-arctic} at both 7B and 32B scales. \textbf{Multi-agent systems} are represented by MARS-SQL\citep{yang2026mars-sql}, which coordinates three 7B models for SQL generation. To ensure fair comparison, all baselines are evaluated using their native inference protocols, avoiding performance degradation from forced harness interactions (see Appendix~\ref{app:harness_cross_eval} for cross-harness ablations).
\vspace{-4pt}
\section{Experimental Results}
\vspace{-3pt}
\subsection{Main Results}
\vspace{-12pt}
\begin{table}[ht]
\caption{Execution accuracy across in-domain (S2-SQLite) and out-of-distribution (BI-mini, LSB-SQLite) benchmarks (S2-SQLite: Spider~2.0-SQLite; BI-mini: BIRD-Interact Mini; LSB-SQLite: LiveSQLBench Base-Lite SQLite). \textbf{Bold} indicates the best, and \underline{underline} indicates the second best.}
\label{tab:main}
\centering
\small
\setlength{\tabcolsep}{4.5pt}
\resizebox{\textwidth}{!}{%
\begin{tabular}{lccccccc}
\toprule
Method & Model Size & Agentic? & Reasoning? & Harness-trained?
& S2-SQLite & BI-mini & LSB-SQLite \\
\midrule
\multicolumn{8}{l}{\textit{General-purpose models}} \\
Qwen2.5-Coder   & 32B          & \xmark  & \checkmark & \xmark & 16.3\%     & 3.0\%     & 7.4\% \\
\midrule
\multicolumn{8}{l}{\textit{SQL-specialized models}} \\
OmniSQL          & 7B           & \xmark     & \xmark     & \xmark & 13.3\% & 1.7\%     & 7.0\% \\
OmniSQL          & 32B          & \xmark     & \xmark     & \xmark & 14.8\% & 2.7\%     & 8.9\% \\
Arctic-Text2SQL  & 7B           & \xmark     & \checkmark & \xmark & 15.6\% & 2.3\%     & 7.0\% \\
Arctic-Text2SQL  & 32B          & \xmark     & \checkmark & \xmark & 16.3\% & --     & -- \\
\midrule
\multicolumn{8}{l}{\textit{Multi-agent systems}} \\
MARS-SQL         & 3$\times$7B  & \checkmark & \checkmark & \xmark & 12.6\% & 3.0\%     & 7.4\% \\
\midrule
\multirow{2}{*}{\method}
& 8B           & \checkmark & \checkmark & \checkmark & \underline{45.2\%} & \underline{9.0\%}  & \underline{24.8\%} \\
& 14B          & \checkmark & \checkmark & \checkmark & \textbf{54.8\%}    & \textbf{11.3\%}    & \textbf{28.5\%} \\
\bottomrule
\end{tabular}%
}
\end{table}

Table~\ref{tab:main} compares our final harness-native agents against existing SQL baselines across in-domain (ID) and out-of-distribution (OOD) benchmarks. Along with execution accuracy, we report model size, agentic interaction, explicit reasoning, and whether each model is trained within an interactive harness, making the differences in model scale and interaction paradigm explicit.

\textbf{In-Domain Performance.} On Spider~2.0-SQLite, \method{} establishes a substantial advantage over both traditional non-agentic baselines and prior multi-agent setups. Our 8B variant achieves 45.2\% execution accuracy, nearly tripling the performance of 32B-scale specialized models such as OmniSQL (14.8\%), Arctic-Text2SQL (16.3\%), and Qwen2.5-Coder-32B (16.3\%). Scaling the policy to 14B yields further consistent gains, reaching 54.8\%. This highlights that compact models, when natively grounded in the execution harness, can vastly outperform far larger open-loop models that lack trajectory-level interaction capabilities.

\textbf{Out-of-Distribution Generalization.} The performance gains transfer robustly to unseen environments and distinct interaction protocols. On BI-mini, which requires multi-turn dialogue with a user simulator, \method{} reaches 9.0\% (8B) and 11.3\% (14B), well above the 1.7\%--3.0\% range achieved by existing 7B--32B baselines and the MARS-SQL multi-agent pipeline (3.0\%). Crucially, on LSB-SQLite, our 8B and 14B models achieve 24.8\% and 28.5\% execution accuracy, respectively, outperforming all evaluated 32B baselines by a large margin (7.0\%--8.9\%). These trends confirm that harness-native post-training instills resilient procedural reasoning and closed-loop error-recovery capabilities rather than simply memorizing in-domain schema patterns.

\subsection{Effect of Post-Training}

Table~\ref{tab:training-ablation} reports the performance progression across different post-training stages on Spider~2.0-SQLite, starting from the original Qwen3 base models.

\textbf{Trajectory Supervision vs. Direct RL.} 
Supervised fine-tuning (SFT) substantially elevates execution accuracy, rising from 15.5\% to 30.4\% for the 8B model and from 22.2\% to 37.8\% for the 14B model (+14.9 and +15.6 percentage points, respectively). 
In contrast, applying execution-reward RL directly to the 8B base model achieves only 20.0\% (+4.5 points). 
This demonstrates that while uninitialized RL provides modest exploration gains, cold-starting via verified harness trajectories is critical for establishing effective task grounding.
\begin{wraptable}{r}{0.34\textwidth}
\vspace{-0.5\baselineskip}
\caption{Effect of harness-native post-training on Spider~2.0-SQLite execution accuracy.}
\label{tab:training-ablation}
\centering
\small
\resizebox{\linewidth}{!}{%
\begin{tabular}{lcc}
\toprule
Training Stage & 8B & 14B \\
\midrule
Base   & 15.5\% & 22.2\% \\
Direct RL & 20.0\% & - \\
SFT  & 30.4\% & 37.8\% \\
SFT + RL   & \textbf{45.2\%} & \textbf{54.8\%} \\
    \bottomrule
    \end{tabular}%
    }
    \vspace{-0.5\baselineskip}
    \end{wraptable}

\textbf{Execution-Reward RL after SFT.} 
Continuing from the SFT checkpoints, RL yields a further substantial leap to 45.2\% (+14.8 points) at 8B and 54.8\% (+17.0 points) at 14B. 
The complete two-stage pipeline delivers total gains of 29.7 and 32.6 points over the respective base models. 
Notably, RL post-SFT yields over $3\times$ the accuracy increment compared to Direct RL on the 8B model, highlighting that trajectory supervision provides the structural foundation necessary for reinforcement learning to efficiently optimize complex execution-guided reasoning.

\vspace{-3pt}
\subsection{Ablation Studies}

We next examine the design choices most closely tied to our formulation: alignment between post-training and the deployment harness, preservation of complete interaction trajectories, and sufficient context for long-horizon interaction. We additionally test the robustness of our results to the RL optimization algorithm. Further analyses of trajectory synthesis and learned interaction behavior are provided in Appendix~\ref{app:additional-ablation}.

\paragraph{Harness-native training and inference.}
\begin{table}[ht]
\vspace{-18pt}
\caption{Comparison of non-interactive vs.\ harness-native training and inference across ID and OOD benchmarks.}
\label{tab:ablation-harness-native}
\centering
\setlength{\tabcolsep}{8pt}
\begin{tabular}{llccc}
\toprule
Post-training & Inference & S2-SQLite (ID) & BI-mini (OOD) & LSB-SQLite (OOD) \\
\midrule
None          & Non-harness & 2.2\%   & 1.0\% & 3.3\%  \\
None          & DSH-SQL     & 15.5\%  & 3.3\% & 12.2\% \\
\midrule
Non-harness   & Non-harness & 14.1\% & 1.3\% & 3.7\%  \\
Non-harness   & DSH-SQL     & 9.6\%  & 3.0\% & 8.2\% \\
Harness-native & DSH-SQL     & 30.4\%  & 5.3\% & 16.6\% \\
\bottomrule
\end{tabular}
\end{table}
We first ablate the core premise of our framework: aligning post-training with the interactive execution harness. Table~\ref{tab:ablation-harness-native} separates the effects of interactive inference and harness-native training.

First, an executable interactive environment is already valuable without post-training: applying DSH-SQL to the base model improves S2-SQLite from 2.2\% to 15.5\%, with consistent gains on both OOD benchmarks.
Second, with inference fixed to DSH-SQL, harness-native post-training substantially outperforms non-harness post-training across all three benchmarks, including 30.4\% vs.\ 9.6\% on S2-SQLite.
Finally, non-harness post-training improves non-harness inference
(2.2\% to 14.1\%) but transfers poorly to DSH-SQL (9.6\%), demonstrating the importance of training--inference alignment. Additional comparisons between the generic Full DSH interface and our SQL-specific harness are reported in Appendix~\ref{app:synthesis-ablation}.

\paragraph{SFT trajectory organization.}
\begin{wrapfigure}{r}{0.5\textwidth}
\vspace{-14pt}
\centering
\includegraphics[width=\linewidth]{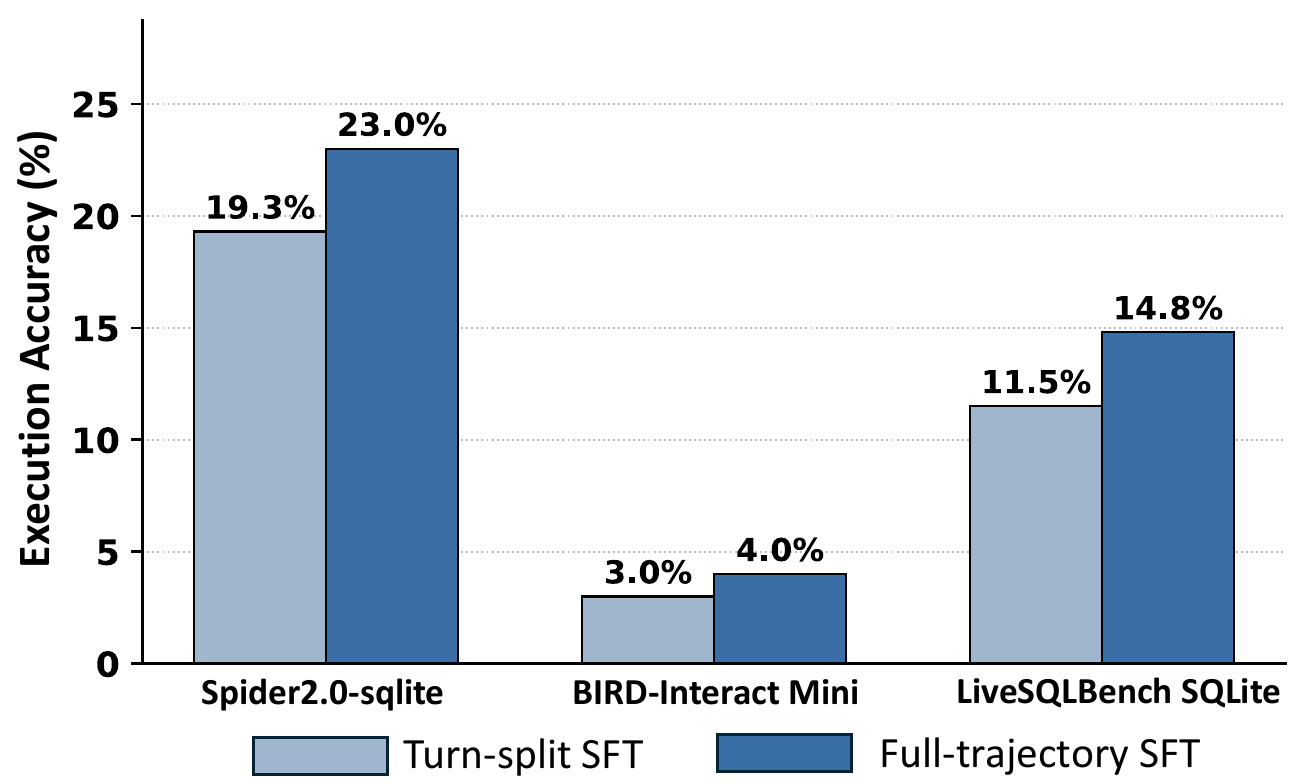}
\vspace{-14pt}
\caption{Effect of multi-turn trajectory organization for SFT.}
\vspace{-6pt}
\label{fig:traj_org_comp}
\end{wrapfigure}
We next investigate whether multi-turn interactions should be trained as unified long sequences or decomposed into turn-level instances. In our default \emph{full-trajectory} setting, an entire episode is trained as a single sequence where all assistant actions are jointly supervised while environment observations are masked from the loss. In contrast, the \emph{turn-split} variant decomposes each episode into separate training samples, optimizing the loss solely on each individual assistant turn conditioned on its preceding history prefix.

As shown in Figure~\ref{fig:traj_org_comp}, full-trajectory SFT consistently outperforms turn-split training across all three benchmarks. This result supports our central supervision hypothesis: preserving complete interaction structure provides a stronger learning signal for long-horizon database agents than decomposing trajectories into isolated decisions.

\paragraph{Trajectory context length.}
\begin{wrapfigure}{r}{0.5\textwidth}
\centering
\includegraphics[width=\linewidth]{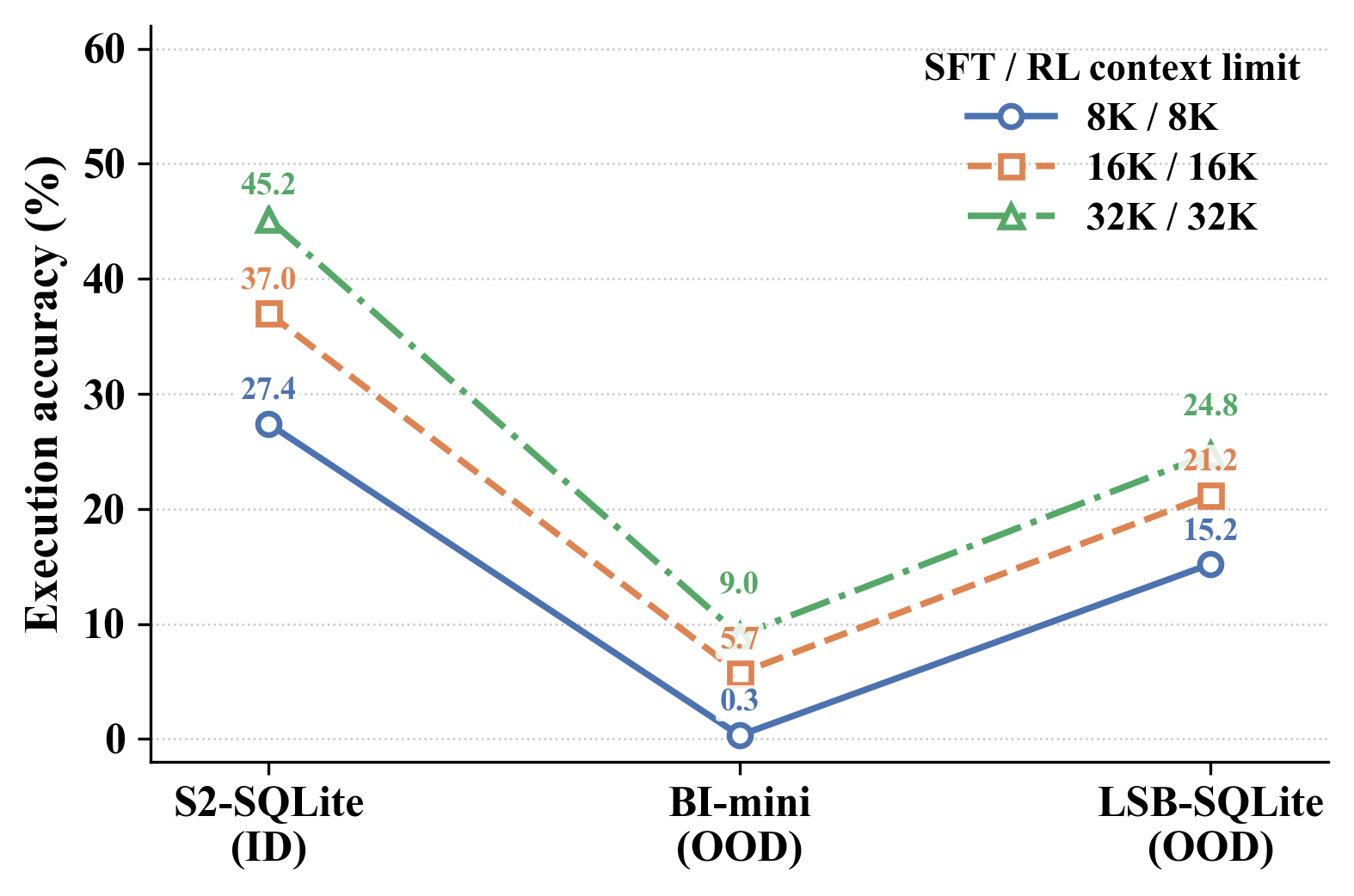}
\caption{Effect of the maximum trajectory context length. The same context limit is used during both SFT and RL.}
\label{fig:context-length}
\vspace{-5pt}
\end{wrapfigure}
Long-horizon SQL agents impose substantially greater context demands than conventional Text-to-SQL systems, as a single trajectory must accommodate reasoning traces, tool calls, execution outputs, intermediate observations, and iterative revisions\citep{pmlr-v306-he26ar}. We therefore vary the maximum sequence length while keeping the context limit consistent across SFT and RL. As shown in Figure~\ref{fig:context-length}, performance improves consistently as the context window increases from 8K to 32K across both ID and OOD benchmarks. The improvement is particularly pronounced on S2-SQLite, where execution accuracy increases from 27.4\% at 8K to 45.2\% at 32K, while BI-mini and LSB-SQLite exhibit similar upward trends. These results suggest that sufficient context capacity is important for learning long-horizon agentic trajectories, where truncating earlier interactions can remove execution evidence and intermediate reasoning required for subsequent decisions.

\paragraph{RL algorithm.}
Our main results use a fixed RL recipe. To determine whether the gains depend strongly on the specific optimization algorithm, we compare GRPO~\citep{shao2024deepseekmath}, GSPO~\citep{zheng2025gspo}, and DAPO~\citep{yu2025dapo} starting from the same SFT checkpoint and using identical rollout data, reward signals, context limits, and training budgets.

\begin{wraptable}{r}{0.55\textwidth}
\vspace{-0.8em}
\centering
\caption{Comparison of RL algorithms under matched training and rollout budgets.}
\label{tab:ablation-rl}
\small
\setlength{\tabcolsep}{5pt}
\resizebox{\linewidth}{!}{%
\begin{tabular}{lccc}
\toprule
RL Algorithm & S2-SQLite & BI-mini & LSB-SQLite \\
\midrule
GRPO & 40.7\% & 8.7\% & 23.3\% \\
GSPO & 43.7\% & 9.0\% & \textbf{25.2\%} \\
\textbf{DAPO} & \textbf{45.2\%} & \textbf{9.0\%} & 24.8\% \\
\bottomrule
\end{tabular}%
}
\vspace{-0.8em}
\end{wraptable}

As shown in Table~\ref{tab:ablation-rl}, both DAPO and GSPO consistently outperform GRPO across both in-domain and out-of-distribution suites, with DAPO achieving the highest accuracy on Spider2.0-SQLite (45.2\% vs. 40.7\%). This improvement is consistent with the more stable policy updates of DAPO and GSPO, which may be particularly beneficial for long-horizon trajectories with sparse execution rewards. The relatively small difference between DAPO and GSPO further suggests that the gains of \method{} are not tied to a single RL algorithm.

Beyond execution accuracy, trajectory-level analysis shows that
harness-native post-training substantially improves protocol adherence, error recovery, and valid termination, providing behavioral evidence that the model learns the intended interaction policy rather than only improving terminal accuracy (Appendix~\ref{app:behavior-analysis}).

\section{Related Work}

\paragraph{LLMs for Text-to-SQL.}
Text-to-SQL has progressed from cross-domain semantic parsing benchmarks such as Spider~\citep{yu2018spider} and BIRD~\citep{li2023bird} to more realistic settings involving large schemas, external knowledge, and enterprise database workflows, as exemplified by Spider2.0\citep{fang2025spider2} and BIRD-Interact~\citep{huo2025bird-interact}.
Recent work improves LLM-based text-to-SQL through large-scale synthetic supervision and execution-based optimization, including OmniSQL and Arctic-Text2SQL-R1~\citep{li2025omnisql,yao-etal-2026-arctic}, while systems such as CHESS~\citep{talaei2024chess}, ReFoRCE~\citep{deng2025reforce}, and MARS-SQL~\citep{yang2026mars-sql} increasingly rely on retrieval, database exploration, execution feedback, and iterative refinement at inference time.
Together, these developments suggest that performance on realistic SQL tasks depends not only on SQL generation ability, but also on effective interaction with the surrounding execution environment.
Rather than proposing another SQL-specific inference strategy, \method{} uses SQL as a challenging domain for studying how training data can be synthesized to teach models to operate effectively through an agent harness.

\paragraph{Harnesses and tool-using agents.}
Modern language-model agents operate through harnesses that define their tools, observations, execution loops, and context management. ReAct~\citep{yao2023react} established the paradigm of interleaving reasoning and actions, while subsequent systems study richer interfaces for API use, executable code, and software environments~\citep{qin2023toolllm,wang2024codeact,yang2024sweagent}. Interactive benchmarks further expose the execution environment as part of the problem~\citep{liu2023agentbench,yao2024taubench}; most directly, Harness-Bench shows substantial performance variation across model–harness pairings even under controlled tasks and validators~\citep{yao2026harnessbench}. These findings motivate treating agent behavior as a property of the model–harness system rather than the backbone alone. \method{} extends this perspective from evaluation to learning by making the deployment harness part of task generation, teacher rollout, post-training, and evaluation.

\paragraph{Synthetic instruction and agent trajectory data.}
Synthetic supervision has evolved from bootstrapped instruction–response pairs~\citep{wang2023selfinstruct,wu2026syntheticlayered} and self-supervised tool-use annotations~\citep{schick2023toolformer} to multi-step tool-use data~\citep{tang2023toolalpaca,qin2023toolllm} and complete agent trajectories~\citep{chen2023fireact,zeng2023agenttuning,mitra2024agentinstruct,he2026clintutor,zheng2027skillprox}. Recent work also emphasizes executable and semantic verification to improve the reliability of synthetic tool-use data~\citep{liu2024apigen,liu2024toolace},as well as step-level confidence calibration for multi-step
  self-correction~\citep{he2025mmboundary}. \method{} combines these directions through \emph{harness-native} synthesis: tasks are grounded in executable environments, solved by teachers through the target harness, and retained only after outcome verification. Task construction, trajectory generation, and verification therefore share the deployment protocol, yielding supervision for the complete harness-conditioned interaction policy rather than isolated answers or tool calls.

\section{Limitations and Ongoing Work}
Our current study focuses primarily on SQLite, while several realistic SQL benchmarks additionally involve dialects such as PostgreSQL, Snowflake, and BigQuery. This limitation is partly inherited from the underlying small models, whose initial capabilities are substantially weaker and less consistent across non-SQLite dialects. Prior work has similarly identified SQL dialect generalization as a major bottleneck for smaller text-to-SQL models, and has shown that execution-verified adaptation can substantially improve performance on previously underrepresented dialects~\citep{zhang2025exesql}. In our setting, this challenge is further compounded by long-horizon interaction, where dialect-specific syntax, database clients, metadata commands, and execution behavior all become part of the agent trajectory. Our ongoing work therefore extends harness-native synthesis to dialect-specific execution environments, with the goal of closing the current performance gap between SQLite and other practically important SQL dialects.


\section{Conclusion}

We presented \method{}, a harness-native framework for training long-horizon SQL agents. Our central finding is that closing the train--deploy gap requires alignment at three levels: an executable environment that exposes authentic database states and runtime feedback, a task-specific harness that defines the agent's interaction protocol, and trajectory-level supervision that teaches the resulting closed-loop policy. \method{} realizes these principles through isolated SQL sandboxes with hidden executable oracles, a minimal four-tool SQL harness, and harness-native full-trajectory SFT followed by execution-reward RL.

Across Spider~2.0-SQLite, BIRD-Interact Mini, and LiveSQLBench Base-Lite SQLite, harness-native post-training consistently improves both tested model scales and transfers to out-of-distribution interactive settings. On Spider~2.0-SQLite, Qwen3-8B improves from 15.5\% to 45.2\% and Qwen3-14B from 22.2\% to 54.8\%. Controlled analyses further show that these gains depend not only on access to an interactive execution environment, but also on aligning training with the deployment harness and preserving complete interaction trajectories during supervision. These results support a broader view of agent learning: the harness is not merely an inference-time wrapper, but part of the behavior distribution that training should reproduce. While broader dialect and harness coverage remains an important direction, our results provide a concrete path toward compact database agents that learn verified interaction policies rather than isolated terminal answers.

\subsection*{Reproducibility Statement}
Appendix~\ref{app:harness} specifies the harness composition and runtime hooks, Appendix~\ref{app:training} specifies the post-training objective and context budgets, Appendix~\ref{app:format} specifies the task and trajectory records, and Appendix~\ref{app:gates} lists the verifier gates used in the current implementation. We record generation prompts, database identifiers, random seeds, raw rollouts, executed SQL, normalized result fingerprints, per-seed outcomes, selected-trajectory provenance, timing, and rejection reasons. The final submission will include an anonymized code snapshot, data manifests, training hyperparameters, environment versions, and evaluation commands; proprietary database contents and model endpoints will be replaced by documented public equivalents where redistribution is not permitted.

\subsection*{AI use statement}
We used an AI assistant for grammar checking and translation during manuscript preparation. For grammar checking, we used the assistant to identify grammatical errors and suggest corrections to sentence structure, tense, and word usage. For translation, we used it to translate text, with attention to fluency, consistent technical terminology, and preservation of the original meaning.
\bibliography{iclr2027_conference}
\bibliographystyle{iclr2027_conference}

\appendix

\section{DSH-SQL Harness Composition and Lifecycle}
\label{app:harness}

\paragraph{System boundary.}
DSH-SQL is a profile-level composition, not a replacement agent loop. Table~\ref{tab:harness-layers} separates inherited DSH services, our SQL-specific additions, and orchestration that runs outside DSH. This distinction is important for attribution: our contribution is the SQL profile, environment adapter, and harness-aligned data and evaluation pipeline; the generic plugin lifecycle, agent loop, and session substrate come from DSH.

\begin{table}[ht]
\caption{Layers of the evaluation harness. ``Visible'' indicates whether the model can invoke or directly observe the component, rather than whether its effects can influence the trajectory.}
\label{tab:harness-layers}
\centering
\small
\setlength{\tabcolsep}{4pt}
\resizebox{\textwidth}{!}{%
\begin{tabular}{p{0.17\textwidth}p{0.18\textwidth}p{0.49\textwidth}c}
\toprule
Layer & Ownership & Responsibility & Visible \\
\midrule
Experiment controller & Ours, outside DSH & Loads tasks; creates a process, workspace, and DSH home per task; sets environment bindings; enforces episode deadlines; schedules workers; recovers submissions; records all outcomes. & No \\
Profile composition & DSH + ours & Composes \texttt{dsh-base}, \texttt{dsh-headless}, and \texttt{dsh-bundle-sql}; disables unrelated coding-agent capabilities; mounts the selected model and database backend. & No \\
Control plane & DSH & Prompt and model routing, agent/session state, durable checkpoints, cancellation, retry, compaction, result retention, and headless termination. & Indirect \\
ReAct-style loop & DSH & Alternates model requests with validated tool dispatch until a terminal response, maximum-token stop, error, or cancellation. & Indirect \\
SQL environment plane & Ours & Exposes backend-specific tools; validates SQL; executes queries; formats bounded observations; implements budget, phase, and submission semantics. & Yes \\
Executable verifier & Ours, outside policy context & Re-executes the submitted artifact, compares normalized results with the hidden oracle, and checks protocol predicates. & No \\
\bottomrule
\end{tabular}%
}
\end{table}

\paragraph{Lifecycle and middleware.}
Cordis provides scoped plugin lifecycle effects and ordered event waterfalls. The former install and remove services and listeners safely; the latter act as middleware around an agent step or tool execution. In simplified order, one step is
\begin{center}
\small
\begin{tabular}{c}
\texttt{system-prompt/assemble} $\rightarrow$ \texttt{agent/pre-step} $\rightarrow$ \texttt{agent/request} \\
$\downarrow$ \\[-0.3em]
\texttt{llm/stream} $\rightarrow$ \texttt{tools/pre-execute} $\rightarrow$ \texttt{tools/execute} \\
$\downarrow$ \\[-0.3em]
SQL tool body $\rightarrow$ \texttt{tools/post-execute} $\rightarrow$ \texttt{tools/result}.
\end{tabular}
\end{center}
The pre-step, model-request, and top-level tool boundaries flush the session before the next external action. The request-error branch supports provider retry and context-overflow recovery. Pre-execution middleware can deny or request approval; execution middleware propagates cancellation and optional tool deadlines; post-execution middleware can bound large outputs and inject corrective context. Every durable event---including prompt headers, token usage, calls, results, compactions, and turn termination---is written to the task's session log.

\paragraph{Composition does not imply activation.}
The inspected SQLite profile contains 82 registrations: 40 are mounted and 42 are disabled. The disabled set includes model-facing shell, filesystem, web, editor, code-runtime, skill, goal, plan, subagent, and workflow capabilities. Among the mounted set, some services are conditional middleware rather than actions that fire in every episode. Table~\ref{tab:harness-activation} reports an implementation audit over one complete 135-task evaluation directory. The directory contains 138 session logs because three tasks were attempted more than once; counts below are over logs rather than unique benchmark items.

\begin{table}[H]
\caption{Runtime activation audit for the SQLite profile. A mounted hook may be traversed without its conditional action firing.}
\label{tab:harness-activation}
\centering
\small
\setlength{\tabcolsep}{3pt}
\begin{tabular}{p{0.39\textwidth}rp{0.42\textwidth}}
\toprule
Mechanism & Observed & Interpretation \\
\midrule
Session policy initialization & 138 & Permission preset, sandbox mode, and approval policy were recorded once per session. \\
\texttt{sql\_list\_tables} / \texttt{sql\_schema} & 153 / 199 & Both tools appeared in every audited session. \\
\texttt{sql\_exec} / \texttt{sql\_submit} & 3,230 / 129 & All sessions explored with execution; 129 reached explicit submission. \\
Automatic context compaction & 22 & The pressure-triggered summarization branch fired only on long trajectories. \\
Tool-result pruning & 1 & One over-threshold observation was deterministically shortened. \\
Repeated-call reminders & 19 & Middleware injected feedback after repeated identical calls. \\
SQLite query deadline & 35 & The adapter killed child queries exceeding 60 seconds. \\
Multi-statement guard & 2 & The SQL adapter rejected calls containing more than one statement. \\
LLM retry / output spill & 0 / 0 & Both plugins were mounted, but their trigger conditions did not occur. \\
Generic DSH tool deadline & 0 & SQL tools use the adapter's child-process deadline rather than a tool-definition deadline. \\
\bottomrule
\end{tabular}
\end{table}

Other mounted base services---for example settings, credentials, attachments, jobs, session querying, and generic shell/sandbox infrastructure---satisfy composition dependencies or initialize ambient state but are not model-callable and are not on the semantic SQL path. We therefore do not count the 40 mounted registrations as 40 agent capabilities.

\paragraph{Backend-specific environment planes.}
The SQL plugin selects one backend per process. SQLite exposes the four tools used for synthesis and Spider evaluation. It checks the single-statement read-only grammar, opens the database read-only, and executes each query in a separately killable child process. PostgreSQL and BigQuery adapters share the four-tool abstraction while implementing dialect-specific execution and scan constraints; they are not used by the reported SQLite results. The BIRD-Interact profile instead exposes the benchmark's nine operations: query execution, schema access, column semantics, three external-knowledge lookups, user clarification, and submission. A wrapper charges the configured bird-coin cost, persists actions and remaining budget, blocks further exploration after exhaustion, and transitions a successful phase-1 submission to the phase-2 user follow-up. The \texttt{ask\_user} operation forwards one question to the GPT-4o user-simulator service. Thus the BIRD experiment preserves DSH's control plane while intentionally changing both the backend and the model-visible capability contract.

\section{Additional Ablations and Analysis}
\label{app:additional-ablation}

\subsection{Trajectory Synthesis Configuration}
\label{app:synthesis-ablation}
We ablate two core design choices in trajectory synthesis: the strength of the teacher model and the specialization of the execution harness. Using Qwen3-8B fine-tuned on a fixed 1000-example synthesized subset, Table~\ref{tab:ablation-synthesis-config} reveals two critical insights:
\begin{table}[h]
\vspace{-10pt}
\caption{Ablation on trajectory synthesis configurations evaluated on Spider~2.0-SQLite, using Qwen3-8B fine-tuned on 1{,}000 synthesized trajectories.}
\label{tab:ablation-synthesis-config}
\centering
\setlength{\tabcolsep}{7pt}
\begin{tabular}{llc}
\toprule
Teacher Model & Harness & S2-SQLite \\
\midrule
\textit{Base Model} & Full DSH & 13.3\% \\
\midrule
Qwen3.5-9B  & DSH-SQL  & 7.4\% \\
Qwen3.6-27B & Full DSH & 15.5\% \\
Qwen3.6-27B & DSH-SQL  & 23.0\% \\
\bottomrule
\end{tabular}
\vspace{-10pt}
\end{table}
First, \textbf{teacher capability directly dictates downstream policy quality.} Weak supervision from Qwen3.5-9B degrades performance below even the un-finetuned base model, confirming that suboptimal trajectories and noisy exploration heavily mislead student learning. High-capacity teachers are therefore indispensable for viable trajectory synthesis.

Second, \textbf{specialized harnesses facilitate both inference and learning for small size models.} Compared to the general-purpose Full DSH, our streamlined DSH-SQL yields superior performance across the board—not only boosting zero-shot base inference (15.5\% vs.\ 13.3\%, as shown previously) but also maximizing post-training gains. Narrowing the action space and reducing interface noise make demonstrations significantly easier for compact models to absorb effectively.

\subsection{Interaction Behavior Analysis}
\label{app:behavior-analysis}
Execution accuracy alone does not reveal whether harness-native post-training actually changes how the model interacts with the database environment. We therefore analyze the complete interaction trajectories of the base, SFT, and SFT+RL models. For each checkpoint, we perform four independent rollouts on all 135 Spider~2.0-SQLite tasks, yielding 540 trajectories per model.
\begin{table}[t]

\caption{Interaction behavior across post-training stages on Spider~2.0-SQLite. }
\label{tab:interaction-behavior}
\centering
\setlength{\tabcolsep}{5pt}
\begin{tabular}{lccc}
\toprule
Metric & Base & SFT & SFT+DAPO \\
\midrule
Trajectories with invalid calls $\downarrow$
    & 67.41\% & 17.41\% & \textbf{0.56\%} \\
SQL execution-error rate $\downarrow$
    & 17.93\% & 9.03\% & \textbf{8.08\%} \\
Immediate recovery $\uparrow$
    & 44.09\% & 67.80\% & \textbf{72.72\%} \\
Final SQL executed before submit $\uparrow$
    & 47.78\% & 70.56\% & \textbf{72.04\%} \\
Exactly one valid successful submit $\uparrow$
    & 18.70\% & 75.74\% & \textbf{93.52\%} \\
Strict protocol violation $\downarrow$
    & 83.15\% & 26.67\% & \textbf{11.48\%} \\
Duplicate-call rate $\downarrow$
    & 58.82\% & 16.29\% & \textbf{6.61\%} \\
\bottomrule
\end{tabular}
\end{table}
Harness-native post-training substantially changes the model's interaction behavior, not only its final execution accuracy. SFT produces the largest improvement in basic protocol adherence: the fraction of trajectories containing invalid calls falls from 67.4\% to 17.4\%, strict protocol violations decrease from 83.2\% to 26.7\%, and duplicate calls drop from 58.8\% to 16.3\%. At the same time, the model becomes substantially more likely to execute its final SQL before submission and to recover immediately from execution errors.

DAPO further consolidates these behaviors. Invalid calls become almost eliminated (0.6\%), exactly-one-valid-submit trajectories increase from 75.7\% after SFT to 93.5\%, and strict protocol violations fall to 11.5\%. In contrast, the reduction in SQL execution errors is more modest (9.0\% to 8.1\%). This suggests a complementary role for the two stages: trajectory SFT teaches the interaction protocol and common recovery patterns, while execution-reward RL primarily improves episode-level behavioral consistency and successful termination.

\section{Post-training Configuration and Objectives}
\label{app:training}

\subsection{Training configuration}

Tables~\ref{tab:sft-training-config} and~\ref{tab:rl-training-config} report the launch configurations of the checkpoints used in our main results. The two model chains are not identical: Qwen3-8B and Qwen3-14B RL are initialized from the 32K SFT checkpoint. Both RL runs allow a 32,768-token DSH session, with at most 4,096 newly generated tokens in any single model call. The trainer's token cap controls dynamic batching for memory use and is not an additional model context-window limit.

\begin{table}[ht]
\centering
\caption{SFT configurations used for both Qwen3-8B and Qwen3-14B checkpoints. Both models share identical training settings and are initialized from their respective base checkpoints. Each example is one complete verified harness episode.}
\label{tab:sft-training-config}
\setlength{\tabcolsep}{8pt}
\begin{tabular}{p{0.32\textwidth}p{0.62\textwidth}}
\toprule
\textbf{Hyperparameter / Setting} & \textbf{Value / Specification} \\
\midrule
Base models & Qwen3-8B / Qwen3-14B \\
Training examples & 2,512 verified trajectories \\
Maximum serialized length & 32,768 tokens \\
Loss mask & Assistant reasoning and actions only; system, user, and environment tokens masked \\
Optimization horizon & 3 epochs (471 optimizer updates) \\
Batching & Per-GPU batch size 1; gradient accumulation steps 2; effective batch size 16 \\
Optimizer & AdamW; $\mathrm{lr}=10^{-5}$; cosine decay; 5\% warmup; weight decay 0.01 \\
Hardware \& parallelization & 8$\times$ NVIDIA H800 GPUs; bfloat16; gradient checkpointing; DeepSpeed ZeRO-3 \\
\bottomrule
\end{tabular}
\end{table}

\begin{table}[ht]
\caption{DAPO configurations used for the reported Qwen3-8B and Qwen3-14B results. ``Prompt pool'' is the number of records in the corresponding launch input.}
\label{tab:rl-training-config}
\centering
\setlength{\tabcolsep}{6pt}
\begin{tabular}{p{0.32\textwidth}p{0.62\textwidth}}
\toprule
Setting & Configuration (Qwen3-8B / Qwen3-14B) \\
\midrule
Initialization & 32K Qwen3-8B / 14B SFT checkpoint \\
Rollout horizon & 150 rollout iterations \\
Rollout group & 4 prompts per iteration; 8 trajectories per prompt \\
Updates and batching & 2 optimizer steps per rollout; global batch 16; micro-batch 1 \\
Context limits & 32,768 tokens per DSH session; at most 4,096 new tokens per model call \\
Optimizer & Adam; $\mathrm{lr}=10^{-6}$; constant schedule; $\beta=(0.9,0.98)$; weight decay 0 \\
Policy objective & DAPO; asymmetric clipping $\epsilon \in [0.20, 0.28]$; no entropy bonus; no effective KL penalty \\
Trainer implementation & 32,768-token dynamic-batch cap; FP32 main gradients \\
Systems & 4 H800 actor GPUs and 4 H800 rollout GPUs; rollout tensor parallelism 4 \\
\bottomrule
\end{tabular}
\end{table}

\subsection{Policy optimization objectives}

\paragraph{GRPO objective.}
For prompt $q$, let $\{\tau_i\}_{i=1}^{G}$ be $G=8$ trajectories with scalar rewards $r_i$. The implementation uses the group-standardized advantage
\begin{equation}
    \hat A_i = \frac{r_i-\operatorname{mean}_{j\in[G]} r_j}
    {\operatorname{std}_{j\in[G]} r_j+10^{-6}},
\end{equation}
and broadcasts $\hat A_i$ to the loss-bearing assistant tokens in trajectory $i$. With token-level importance ratio $\rho_{i,t}=\pi_\theta(a_{i,t}\mid s_{i,t})/\pi_{\mathrm{old}}(a_{i,t}\mid s_{i,t})$, we minimize the clipped policy loss
\begin{equation}
    \mathcal{L}_{\mathrm{GRPO}}(\theta) =
    -\mathbb{E}_{i,t}\!\left[
    \min\!\left(\rho_{i,t}\hat A_i,
    \operatorname{clip}(\rho_{i,t},1-\epsilon,1+\epsilon)\hat A_i\right)
    \right], \qquad \epsilon=0.2.
\end{equation}
The number of optimizer steps and effective global batch size for each checkpoint are reported in Table~\ref{tab:rl-training-config}. Both runs use no entropy bonus. The launch configuration sets the auxiliary KL-loss coefficient to zero. It also passes a reference-KL bookkeeping coefficient of 0.01; in the inspected Slime revision, the standard GRPO return broadcaster does not subtract that computed KL tensor. Consequently, the effective objective for these checkpoints is the clipped, group-normalized execution-reward objective above, without a KL term. Failed harness processes are removed groupwise before optimization rather than treated as negative demonstrations.

\paragraph{GSPO objective.}
Group Sequence Policy Optimization (GSPO) replaces GRPO's token-level importance weighting and clipping with a single sequence-level decision~\citep{zheng2025gspo}. Let $\mathcal{T}_i$ be the loss-bearing assistant-token positions in trajectory $\tau_i$ and $N_i=|\mathcal{T}_i|$. Environment observations remain part of each conditioning state $s_{i,t}$ but do not bear loss. From the token-level ratios above, GSPO constructs the length-normalized trajectory ratio
\begin{equation}
    \sigma_i(\theta)
    = \left(\prod_{t\in\mathcal{T}_i}\rho_{i,t}\right)^{1/N_i}
    = \exp\!\left(\frac{1}{N_i}\sum_{t\in\mathcal{T}_i}
      \log \rho_{i,t}\right).
\end{equation}
Using the same group-standardized trajectory advantage $\hat A_i$, its clipped loss is
\begin{equation}
    \mathcal{L}_{\mathrm{GSPO}}(\theta)
    =-\mathbb{E}_{i}\!\left[
      \min\!\left(\sigma_i(\theta)\hat A_i,
      \operatorname{clip}\!\left(\sigma_i(\theta),
      1-\epsilon_s,1+\epsilon_s\right)\hat A_i\right)
    \right],
\end{equation}
where $\epsilon_s$ is the sequence-level clipping radius. Thus every assistant token in a trajectory is weighted by the same trajectory ratio, and the entire trajectory is clipped together. The length normalization keeps ratios from trajectories of different lengths on a comparable numerical scale. This makes the unit of importance correction and clipping coincide with the unit of our terminal execution reward, whereas GRPO clips each token ratio separately.

\paragraph{DAPO objective.}
Decoupled Clip and Dynamic sAmpling Policy Optimization (DAPO) retains token-level importance ratios but modifies group policy optimization in three relevant ways: it uses different lower and upper clipping radii, removes zero-variance reward groups through dynamic sampling, and normalizes the loss across all loss-bearing tokens in the batch rather than first assigning equal weight to every trajectory~\citep{yu2025dapo}. For binary execution rewards, define the effective prompt set
\begin{equation}
    \mathcal{D}_{\mathrm{eff}}
    =\left\{q:\ 0<\sum_{i=1}^{G}\mathbb{I}[r_i=1]<G\right\}.
\end{equation}
Groups outside this set have $\hat A_i=0$ for every trajectory and hence provide no policy-gradient signal. With $N_i=|\mathcal{T}_i|$, the DAPO loss is
\begin{equation}
\begin{split}
    \mathcal{L}_{\mathrm{DAPO}}(\theta)
    =-&\mathbb{E}_{q\in\mathcal{D}_{\mathrm{eff}}}\!\left[
    \frac{1}{\sum_{i=1}^{G}N_i}
    \sum_{i=1}^{G}\sum_{t\in\mathcal{T}_i}
    \min\!\left(
      \rho_{i,t}\hat A_i,\right.\right.\\[-2pt]
    &\left.\left.\qquad
      \operatorname{clip}\!\left(\rho_{i,t},
      1-\epsilon_{\mathrm{low}},1+\epsilon_{\mathrm{high}}\right)
      \hat A_i
    \right)\right],
\end{split}
\end{equation}
where DAPO typically chooses $\epsilon_{\mathrm{high}}>\epsilon_{\mathrm{low}}$ to permit larger probability increases for positively advantaged, initially unlikely tokens. In the matched ablation, the algorithms receive the same rollout pool before their algorithm-specific reduction or filtering, and DAPO uses the same terminal execution reward and context cutoff as GRPO and GSPO; we do not add a separate overlength shaping reward. This isolates the policy-optimization rule while preserving the common harness interaction budget.

\paragraph{Execution reward.}
Let $E(y)$ execute the submitted SQL $y$ with a 30-second timeout and return its normalized relation. For tasks whose output ordering is semantically required, normalization preserves row order; otherwise, it compares an order-invariant result fingerprint. The terminal reward is
\begin{equation}
R(\tau)=\begin{cases}
1, & E(y)=E(y^*)\text{ under the task's ordered/unordered comparison},\\
1, & \text{values match and the expected aliases are not specified publicly},\\
0, & \text{otherwise}.
\end{cases}
\end{equation}
The zero-reward cases include no parsable final SQL, execution errors or timeouts, wrong values, and wrong public column names. There is no partial credit for syntactic validity, number of tool calls, or intermediate SQL. This keeps the RL signal aligned with executable task success while leaving exploration behavior to be learned through credit assignment over the complete DSH trajectory.

\section{SFT Corpus Statistics}
\label{app:sft-corpus}

Table~\ref{tab:sft-corpus-statistics} summarizes the final SFT corpus after teacher rollout and oracle-based selection. The corpus contains 2,512 successful trajectories collected from 79 database instances across Spider2.0-Lite and Spider2.0-DBT. Each example retains the complete harness interaction rather than only the final SQL query. The resulting trajectories are relatively long, averaging 11.8K tokens and a median of 16 agent turns, with 17.0 tool calls and 13.9 SQL executions per trajectory on average.

The interaction statistics show that successful demonstrations frequently contain non-trivial intermediate behavior. In particular, 77.83\% of trajectories contain at least one failed intermediate SQL execution, while 99.92\% execute at least two distinct SQL queries. Under the stricter criterion of revising the query after an execution failure, 77.71\% of trajectories exhibit such behavior. Thus, the retained demonstrations capture not only successful final queries but also iterative execution, feedback, and revision within the harness.

\begin{table}[t]
\caption{Statistics of the SFT corpus after teacher rollout and oracle-based selection.}
\label{tab:sft-corpus-statistics}
\centering
\setlength{\tabcolsep}{7pt}
\begin{tabular}{lr}
\toprule
Statistic & Value \\
\midrule
Tasks / trajectories & 2,512 \\
Database instances & 79 \\
Average trajectory tokens & 11,793.89 \\
Median agent turns & 16 \\
Average tool calls & 17.00 \\
Average SQL executions & 13.91 \\
Trajectories with $\geq 1$ failed intermediate query & 77.83\% \\
Trajectories with query revision & 99.92\% \\
Trajectories with post-failure query revision & 77.71\% \\
\midrule
Teacher pass@1 & 71.20\% \\
Teacher pass@3 & 83.61\% \\
\bottomrule
\end{tabular}
\end{table}

Across the retained trajectories, we record 42,704 tool calls and 34,936 SQL executions, of which 3,989 return execution failures. These failures should not be interpreted directly as SQL reasoning errors. Of the 3,989 failed executions, 3,230 arise from queries beginning with SQL comments (\texttt{-}), which are rejected by the harness read-only guard because the first parsed token is not \texttt{SELECT}, \texttt{WITH}, or \texttt{EXPLAIN}. We retain these interactions as collected rather than sanitizing them from otherwise successful trajectories, since they reflect the authentic execution feedback observed by the teacher during data collection.

\paragraph{Ablation on Sanitizing Comment-Induced Execution Failures.}
To verify whether exposing the model to such syntactically benign execution rejections during training introduces harmful artifacts, we perform a controlled ablation study. Specifically, we construct a sanitized training set by pruning all failed interaction turns caused solely by leading comment rejections from the original trajectories, and train an identical model checkpoint using the same training configuration. 

When evaluated on the Spider~2.0-Lite SQLite benchmark:
\begin{itemize}
    \item The model trained on the original, unpruned trajectories achieves an execution accuracy of \textbf{30.4\%}.
    \item The model trained on the sanitized trajectories achieves an execution accuracy of \textbf{28.2\%} (a drop of 2.2 percentage points).
\end{itemize}

This performance gap demonstrates that retaining benign execution failures along with the subsequent recovery turns provides valuable supervised signal for error recovery and environment self-correction. Sanitizing these interaction cycles reduces the diversity of feedback trajectories, making the model more brittle when encountering real-world runtime rejections from the evaluation harness.

\section{Data Integrity and Leakage Audit for Shared-Database Evaluation}
\label{app:leakage_audit}

In our main experiments, the agent is fine-tuned on synthetic tasks derived from the 30 SQLite database instances within Spider~2.0-Lite and evaluated on the official 135 Spider~2.0-Lite SQLite test tasks. This paradigm reflects a realistic enterprise setting where database schemas remain stable while queries vary over time. To ensure that performance gains stem from procedural task execution rather than rote task memorization, we conduct a rigorous multi-tier audit confirming that the evaluation tasks were not present in or leaked into the synthetic training corpus.

Because the Spider~2.0 evaluation benchmark assesses execution outcomes via accepted ground-truth CSV outputs rather than canonical Gold SQL queries, we audit orthogonality across three complementary dimensions: lexical similarity, chunk-level semantic embedding, and database execution results.

\subsection{Exact Matching and Near-Duplicate Analysis}

We first perform exact matching and lexical overlap checks between the 135 SQLite test queries and both the full synthetic generation pool and the final SFT subset. As summarized in Table~\ref{tab:leakage_summary}, across all 30 byte-identical SQLite database instances (verified via SHA-256), there is strictly zero overlap in Task IDs, exact question strings, normalized question strings, and substring containment.

\begin{table}[h]
\centering
\resizebox{\columnwidth}{!}{%
\begin{tabular}{lcc}
\toprule
\textbf{Audit Check} & \textbf{Full Synthetic Pool vs. Spider2.0} & \textbf{Final SFT Subset vs. Spider2.0} \\
\midrule
Database Names & 30 / 30 Identical & 30 / 30 Identical \\
Database File SHA-256 & 30 / 30 Byte-identical & 30 / 30 Byte-identical \\
Task ID Intersection & 0 & 0 \\
Exact Question Matches & 0 & 0 \\
Normalized Exact Matches & 0 & 0 \\
Same-DB Question Containment & 0 & 0 \\
Same-DB Execution Matches (Unordered) & 0 & 0 \\
Semantic Embedding Cosine $\ge 0.75$ & --- & 0 / 135 \\
\bottomrule
\end{tabular}
}
\caption{Comprehensive leakage and overlap audit between synthetic training data and Spider~2.0-Lite SQLite-135 evaluation tasks.}
\label{tab:leakage_summary}
\end{table}

To rule out subtle surface paraphrasing, we measure nearest-neighbor distance over normalized word tokens across all pairs within the same database:
\begin{itemize}
    \item \textbf{Token Sequence Ratio:} Across all 135 test queries, the median sequence ratio to its nearest same-DB synthetic counterpart is 0.0913 (P95: 0.1352; Maximum: \textbf{0.1689}).
    \item \textbf{Token-set Jaccard Similarity:} The median Jaccard index is 0.0971 (P95: 0.1600; Maximum: \textbf{0.1869}).
    \item \textbf{Longest Common Contiguous Span:} The median contiguous token span is only 2 tokens (P95: 4; Maximum: \textbf{5 tokens}).
\end{itemize}

None of the test queries remotely approach standard duplicate thresholds (e.g., $0.5$ Jaccard or sequence ratio). The maximum contiguous span of 5 tokens reflects shared database entities (e.g., column names and domain terminology) rather than replicated phrasing. Manual inspection of the top nearest pairs (detailed in Table~\ref{tab:nearest_pairs}) confirms that high-overlap token pairs define fundamentally distinct business objectives and analytical cohorts.

\begin{table}[h]
\centering
\resizebox{\textwidth}{!}{%
\begin{tabular}{lllc p{7.2cm}}
\toprule
\textbf{Test ID} & \textbf{Nearest Synthetic ID} & \textbf{Database} & \textbf{Seq. Ratio} & \textbf{Task Divergence (Manual Verification)} \\
\midrule
\texttt{local007} & \texttt{local\_syn\_000323} & Baseball & 0.1689 & \textbf{Test:} Lifetime career span across all players. \newline \textbf{Train:} Career home-run metrics for the 1990s debut cohort. \\
\addlinespace[3pt]
\texttt{local310} & \texttt{local\_syn\_000250} & F1 & 0.1688 & \textbf{Test:} Annual minimal combinations of driver and constructor. \newline \textbf{Train:} Constructor points share over 2010--2020. \\
\addlinespace[3pt]
\texttt{local259} & \texttt{local\_syn\_000331} & IPL & 0.1425 & \textbf{Test:} Lifetime batting and bowling profiles per player. \newline \textbf{Train:} Venue-level toss-to-field match aggregations. \\
\addlinespace[3pt]
\texttt{local055} & \texttt{local\_syn\_000045} & Chinook & 0.1419 & \textbf{Test:} Customer spending by top vs. bottom sales artists. \newline \textbf{Train:} Invoice revenue aggregations by billing country. \\
\addlinespace[3pt]
\texttt{local114} & \texttt{local\_syn\_000369} & Education & 0.1412 & \textbf{Test:} Regional sales-representative performance report. \newline \textbf{Train:} FY2021 product-segment revenue breakdown. \\
\bottomrule
\end{tabular}
}
\caption{Manual inspection of top-5 nearest train-test pairs by lexical sequence ratio within shared databases.}
\label{tab:nearest_pairs}
\end{table}

\subsection{Semantic Overlap Analysis}

To capture semantic alignment beyond surface-level n-grams, we compute dense representations using \texttt{sentence-transformers/all-MiniLM-L6-v2} (384-dimensional). To avoid truncation artifacts on longer synthetic task instructions, each text is chunked into overlapping windows of at most 254 tokens with a 48-token stride. We evaluate both the normalized chunk-mean document cosine similarity and the maximum pairwise chunk cosine similarity against all 1,797 same-database synthetic training candidates.

As shown in Table~\ref{tab:semantic_overlap}, the distribution of semantic similarities remains low to moderate:
\begin{itemize}
    \item The median document cosine similarity across the 135 test queries is \textbf{0.4965} (P90: 0.6229; Maximum: \textbf{0.7122}).
    \item Crucially, \textbf{zero} test queries have a document cosine similarity or max-chunk cosine similarity $\ge 0.75$ against any same-database synthetic training candidate.
\end{itemize}

\begin{table}[h]
\centering
\small
\begin{tabular}{lcccccc}
\toprule
\textbf{Metric (over 135 test tasks)} & \textbf{Median} & \textbf{P90} & \textbf{P95} & \textbf{Maximum} & \textbf{$\ge$ 0.75} & \textbf{$\ge$ 0.80} \\
\midrule
Document Cosine Similarity & 0.4965 & 0.6229 & 0.6487 & \textbf{0.7122} & \textbf{0} & \textbf{0} \\
Max-Chunk Cosine Similarity & 0.5241 & 0.6306 & 0.6639 & \textbf{0.7075} & \textbf{0} & \textbf{0} \\
\bottomrule
\end{tabular}
\caption{Semantic similarity distributions between evaluation queries and candidate training queries within the same database.}
\label{tab:semantic_overlap}
\end{table}

Only a single pair reaches a cosine similarity above 0.70: \texttt{local259} vs. \texttt{local\_syn\_000531} (0.7122, on the IPL database). Manual inspection reveals that while both queries involve cricket player terminology, \texttt{local259} requires calculating complete career batting and bowling profiles, whereas \texttt{local\_syn\_000531} constructs a filtered cohort of bowlers who conceded zero sixes in Season~5 and computes country-level runs-conceded metrics in Season $\ge 6$. The observed similarity stems naturally from domain ontology (teams, bowlers, matches) rather than task objective alignment.

\subsection{Execution Result Fingerprinting}

Because lexical and semantic distances alone cannot completely rule out deep semantic paraphrasing, we implement an execution result fingerprinting audit. Since the underlying SQLite databases are identical bit-for-bit, any synthesized query that unintentionally mirrors a test task's analytical intent would produce identical execution outputs on that database.

We compare the reference execution results of all synthetic tasks against the complete set of 328 official accepted ground-truth CSV outputs across the 135 test tasks:
\begin{itemize}
    \item We cross-check every candidate pair within the same database ($N = 14{,}266$ pairs for the full generation pool; $N = 14{,}260$ pairs for the final SFT corpus).
    \item We evaluate across four fingerprint criteria: ordered vs. unordered row sets, and full schema (with column headers) vs. value-only sets.
    \item Cell values undergo standard normalization (handling of NULLs, floating-point precision, whitespace, and case insensitivity).
\end{itemize}

Across all evaluated pairs under the most permissive criterion (\textbf{ignoring column names, ignoring row order, and matching raw result values only}), the number of matches is \textbf{strictly zero}. 

In summary, the 135 test tasks feature zero intersection in task identifiers, query phrasing, semantic embeddings, and execution outputs. These results demonstrate that the observed downstream execution performance cannot be attributed to training data leakage or target memorization.

\section{Cross-Harness Evaluation}
\label{app:harness_cross_eval}

The main comparison in Table~\ref{tab:main} evaluates existing baselines under their native or author-recommended inference protocols rather than forcing all models into DSH-SQL. This is important because these systems are trained for different interaction paradigms: conventional text-to-SQL models directly generate SQL from the input context, whereas agentic models interact with the database through multi-turn tool use and execution feedback. Evaluating a model under an unfamiliar harness therefore introduces an additional inference-protocol mismatch beyond differences in model capability.

To quantify this effect, we evaluate representative baselines under both their native protocols and DSH-SQL, while keeping the benchmark instances and execution-based evaluation unchanged. Under DSH-SQL, the models use the same interactive environment as our agents but receive no additional harness-specific training or adaptation.

\begin{table}[t]
\caption{Cross-harness evaluation of representative baselines. \textit{Native} denotes the original or author-recommended inference protocol; \textit{DSH-SQL} evaluates the same model in our interactive harness without additional harness-specific training.}
\label{tab:cross-harness}
\centering
\small
\setlength{\tabcolsep}{6pt}
\begin{tabular}{llccc}
\toprule
Model & Protocol & S2-SQLite & BI-mini & LSB-SQLite \\
\midrule
\multirow{2}{*}{Qwen2.5-Coder-32B}
& Native  & 16.3\% & 3.0\% & 7.4\% \\
& DSH-SQL & 3.0\% & 0\% & 5.6\% \\
\midrule
\multirow{2}{*}{OmniSQL-32B}
& Native  & 14.8\% & 2.7\% & 8.9\% \\
& DSH-SQL & 1.5\% & 0\% & 0\% \\
\midrule
\multirow{2}{*}{Arctic-Text2SQL-7B}
& Native  & 15.6\% & 2.3\% & 7.0\% \\
& DSH-SQL & 0\% & 0\% & 1.1\% \\
\bottomrule
\end{tabular}
\end{table}

As shown in Table~\ref{tab:cross-harness}, directly transferring these baselines to DSH-SQL consistently degrades performance. The effect is particularly pronounced for SQL-specialized models: OmniSQL-32B drops from 14.8\% to 1.5\% on S2-SQLite, while Arctic-Text2SQL-7B drops from 15.6\% to 0\%. Even the general-purpose Qwen2.5-Coder-32B declines substantially under the unfamiliar interaction protocol. These results indicate that access to an interactive harness alone does not guarantee better performance; models must also be able to operate effectively under its interaction protocol.

This finding motivates our use of native-protocol results for existing baselines in Table~\ref{tab:main}: forcing non-harness-trained models into DSH-SQL would conflate model capability with harness compatibility and systematically disadvantage methods designed for direct SQL generation. Conversely, our controlled ablations in Table~\ref{tab:ablation-harness-native} evaluate training strategies under the same DSH-SQL inference protocol, allowing the contribution of harness-native post-training to be isolated.

\section{Evaluation Benchmarks}
\label{app:benchmarks}

To comprehensively evaluate reasoning capabilities, interactive workflow compliance, and cross-distribution generalization, we evaluate all models across three challenging SQLite-based benchmarks. 

\paragraph{Spider~2.0-SQLite (S2-SQLite).}
Derived from the SQLite subset of Spider~2.0-Lite~\citep{fang2025spider2}, S2-SQLite serves as our primary in-domain (ID) evaluation benchmark containing 135 complex database instances. Unlike traditional single-turn Text-to-SQL tasks with simplified schemas, S2-SQLite reflects authentic software engineering environments: databases feature complex schemas, inter-table constraints, messy multi-table joins, and enterprise data representations. Models must explore tables iteratively, formulate hypotheses, verify intermediate results, and generate the correct query without human intervention.

\paragraph{BIRD-Interact Mini (BI-mini).}
To assess interactive task completion and dialogue-driven requirement disambiguation, we evaluate on the \textit{Mini-Interact} split of BIRD-Interact~\citep{huo2025bird-interact}. We adopt the \textbf{A-Interact} protocol, wherein the model acts as an interactive SQL agent communicating with a simulated human user (powered by GPT-4o). The user simulator provides ambiguous or partial initial requests, answers clarify questions, and gives iterative feedback based on intermediate execution failures. Models are assessed on their ability to minimize unnecessary turns while successfully converging on the gold execution output.

\paragraph{LiveSQLBench Base-Lite SQLite (LSB-SQLite).}
To evaluate zero-shot transfer and robustness against contamination, we evaluate on the SQLite subset of LiveSQLBench~\citep{livesqlbench2026}. LiveSQLBench features continually updated real-world database tasks sourced from open-source repositories and production data pipelines. The Base-Lite partition reflects realistic developer interactions, testing whether models can generalize their multi-turn exploration and schema navigation strategies to unseen database topologies, dirty schemas, and dialect-specific edge cases.

\paragraph{Evaluation Protocol and Tool Contract.}
All models—including baselines and \textsc{HarnessSynth}—are integrated and evaluated uniformly via the \textsc{DSH-SQL} interface. Across all benchmarks, execution accuracy (EX) serves as the primary metric:
\begin{equation}
\text{EX} = \frac{1}{|D|} \sum_{i=1}^{|D|} \mathbb{I}\left( \text{Exec}(Y_i, \mathcal{D}_i) = \text{Exec}(Y^*_i, \mathcal{D}_i) \right),
\end{equation}
where $\text{Exec}(Y, \mathcal{D})$ denotes the result table returned by executing query $Y$ on database $\mathcal{D}$. Missing answers, unparseable outputs, syntax errors, or queries exceeding the execution deadline are strictly materialized as non-matching sentinel errors, counting directly against the denominator.

\section{Data Record and Export Format}
\label{app:format}

Each task blueprint contains at least \texttt{sample\_id}, \texttt{database\_id}, \texttt{template\_id}, instruction, oracle SQL, expected columns, row bounds, operators, semantic roles, joins, and mutations. The environment record adds dialect, connection metadata, tool definitions, permissions, and budgets. The trajectory record stores the full message stream, tool arguments, observations, final SQL, execution result, verifier outcome, and selection method. Training exports remove hidden oracle fields from the model-visible context and mask loss on system, user, and tool-observation tokens.


\section{Structural Difficulty Control and Synthesized Task Examples}
\label{app:difficulty-control}

\subsection{Controllable structural difficulty}

For a synthesized candidate query $z$, let $n_{\mathrm{tok}}$ be its lexical token count; let $n_{\mathrm{cte}}$, $n_{\mathrm{join}}$, $n_{\mathrm{sel}}$, $n_{\mathrm{win}}$, $n_{\mathrm{set}}$, $n_{\mathrm{exists}}$, $n_{\mathrm{cagg}}$, $n_{\mathrm{case}}$, $n_{\mathrm{time}}$, and $n_{\mathrm{json}}$ count the corresponding SQL constructs; and let $d_{\mathrm{sel}}$ be the maximum nesting depth at a \texttt{SELECT}. We compute
\begin{align}
D_{\mathrm{str}}(z)={}&0.012n_{\mathrm{tok}}+1.2n_{\mathrm{cte}}+0.8n_{\mathrm{join}}
+1.4[n_{\mathrm{sel}}-1]_{+} \nonumber\\
&+2.2n_{\mathrm{win}}+2.0n_{\mathrm{set}}+1.5n_{\mathrm{exists}}
+1.2n_{\mathrm{cagg}} \nonumber\\
&+0.8n_{\mathrm{case}}+1.0n_{\mathrm{time}}+1.0n_{\mathrm{json}}
+1.2d_{\mathrm{sel}}.
\label{eq:structural-difficulty}
\end{align}
This score is a transparent structural control variable, not a claim that SQL syntax alone perfectly predicts model success. A generation run specifies a mixture $\pi_b$ over four bounded bands $b\in\{\text{foundation},\text{core},\text{growth},\text{stretch}\}$. Each band provides lower and upper score and token bounds together with a minimum number of advanced structure families:
\begin{equation}
L_b^D\le D_{\mathrm{str}}(z)\le U_b^D,\qquad
L_b^T\le n_{\mathrm{tok}}(z)\le U_b^T,\qquad
|\mathcal{F}(z)|\ge m_b,
\label{eq:difficulty-contract}
\end{equation}
where $\mathcal{F}(z)$ includes measurable families such as multi-stage CTEs, windows, set or anti-joins, nested subqueries, temporal sequences, and nontrivial derived classifications. Changing $\pi_b$ or the band bounds controls the synthesized curriculum without changing the underlying databases. Candidates outside the requested interval receive feature-level feedback and are simplified or enriched through bounded repair; upper bounds prevent uncontrolled complexity escalation. The score and gate are computed only on candidate SQL produced by the synthesis pipeline. No benchmark reference-query text is supplied to either component.

\subsection{Representative synthesized task records}
\label{app:task-examples}

We show two tasks retained in the 2,512-trajectory SFT corpus. The instruction is visible to the rollout policy, whereas the oracle SQL, result fingerprint, and mutations are hidden during rollout. We expose the oracle SQL here only to make the construction auditable. The stored records additionally contain the complete catalog-grounding trace, execution plan, result preview, hashes, and selected teacher trajectory.

\paragraph{Example 1: core task.}
\textbf{Record.} \texttt{sqlite\_candidate\_000002} uses the \texttt{Airlines} database. Its synthesized SQL has $D_{\mathrm{str}}=15.732$ and belongs to the core band. It composes two CTEs, two joins, JSON extraction, conditional aggregation, distinct counting, and a ranking window. The verifier records six output columns, eight result rows, and five result-changing mutations.

\textbf{Public instruction (abridged for presentation).} Profile August 2017 ticket revenue by aircraft model. Keep flights whose status is exactly \texttt{Arrived} and whose scheduled-departure date prefix lies between August 1 and 31, then identify each model by the English value in the aircraft-model JSON. For each model, report the distinct arrived-flight count, total revenue, revenue per flight, business-fare revenue percentage, and rank by raw total revenue. Compute decisions from raw values, round only displayed revenue and ratio fields to two decimals, and order by raw revenue descending with model name as the tie-breaker.

\textbf{Hidden oracle SQL.}
\begingroup
\scriptsize
\begin{verbatim}
WITH arrived_legs AS (
  SELECT f.flight_id,
         json_extract(a.model, '$.en') AS model_en
  FROM flights AS f
  JOIN aircrafts_data AS a
    ON a.aircraft_code = f.aircraft_code
  WHERE f.status = 'Arrived'
    AND substr(f.scheduled_departure, 1, 10)
        BETWEEN '2017-08-01' AND '2017-08-31'
),
model_revenue AS (
  SELECT al.model_en,
         COUNT(DISTINCT al.flight_id) AS flight_ct,
         SUM(tf.amount) AS total_amt,
         SUM(CASE WHEN tf.fare_conditions = 'Business'
                  THEN tf.amount ELSE 0 END) AS business_amt
  FROM arrived_legs AS al
  JOIN ticket_flights AS tf ON tf.flight_id = al.flight_id
  GROUP BY al.model_en
)
SELECT model_en AS aircraft_model,
       flight_ct AS arrived_flights,
       ROUND(total_amt, 2) AS total_revenue,
       ROUND(total_amt * 1.0 / flight_ct, 2) AS revenue_per_flight,
       ROUND(100.0 * business_amt / total_amt, 2)
         AS business_revenue_pct,
       RANK() OVER (ORDER BY total_amt DESC) AS revenue_rank
FROM model_revenue
ORDER BY total_amt DESC, aircraft_model ASC;
\end{verbatim}
\endgroup
\begingroup\raggedright
The first verified row is \texttt{(Sukhoi Superjet-100, 1748, 1603407500.0, 917281.18, 29.58, 1)}.\par
\endgroup

\paragraph{Example 2: stretch task.}
\textbf{Record.} \texttt{dbt\_score20\_30\_000001} uses the \texttt{activity} database. It was generated under the bounded contract $20\le D_{\mathrm{str}}\le30$ and has $D_{\mathrm{str}}=20.824$, 302 SQL tokens, and two measured advanced families: window computation and nontrivial derived classification. The verifier records eight output columns, eight result rows, and five result-changing mutations.

\textbf{Public instruction (abridged for presentation).} Audit how completely an expected event set covers the raw \texttt{input\_\_aggregate\_in\_between} stream. Keep all stream rows and match coverage by \texttt{activity\_id} with a left join. For each entity, anchor the sequence at its earliest \texttt{visit page}; then aggregate by entity and activity, reporting total events, post-anchor events, covered events, covered revenue, within-group coverage, and share of the entity's events. Use raw values for all decisions and round only the final percentages.

\textbf{Hidden oracle SQL.}
\begingroup
\scriptsize
\begin{verbatim}
WITH stream AS (
  SELECT s.activity_id, s.entity_uuid, s.activity, s.ts,
    MIN(CASE WHEN s.activity = 'visit page' THEN s.ts END)
      OVER (PARTITION BY s.entity_uuid) AS first_visit_ts,
    COUNT(*) OVER (PARTITION BY s.entity_uuid) AS entity_event_total
  FROM input__aggregate_in_between AS s
),
matched AS (
  SELECT st.entity_uuid, st.activity, st.ts, st.first_visit_ts,
    st.entity_event_total,
    CASE WHEN d.activity_id IS NULL THEN 0 ELSE 1 END AS is_covered,
    CASE WHEN d.activity_id IS NULL THEN 0
         ELSE COALESCE(
           d.aggregate_in_between_bought_something_revenue_impact, 0)
    END AS covered_revenue
  FROM stream AS st
  LEFT JOIN dataset__aggregate_in_between_1 AS d
    ON d.activity_id = st.activity_id
)
SELECT m.entity_uuid, m.activity,
  COUNT(*) AS events_total,
  SUM(CASE WHEN m.first_visit_ts IS NOT NULL
            AND m.ts >= m.first_visit_ts THEN 1 ELSE 0 END)
    AS events_on_or_after_first_visit,
  SUM(m.is_covered) AS events_in_expected_output,
  SUM(m.covered_revenue) AS expected_bought_revenue,
  ROUND(100.0 * SUM(m.is_covered) / COUNT(*), 2)
    AS pct_events_in_expected_output,
  ROUND(100.0 * COUNT(*) / MAX(m.entity_event_total), 2)
    AS pct_of_entity_events
FROM matched AS m
GROUP BY m.entity_uuid, m.activity
ORDER BY m.entity_uuid ASC, COUNT(*) DESC, m.activity ASC;
\end{verbatim}
\endgroup
The first verified row is \texttt{(1, visit page, 4, 4, 4, 100, 100.0, 50.0)}. Together, the core and stretch examples illustrate how changing the bounded difficulty contract alters the required composition depth while preserving the same catalog-grounded, execute-and-verify construction protocol.

\section{Verifier Checklist}
\label{app:gates}

\begin{enumerate}
    \item \textbf{Static SQL safety:} a single read-only \texttt{SELECT} or \texttt{WITH} statement; no write, DDL, attachment, or multi-statement escape.
    \item \textbf{Schema validity:} referenced tables and columns exist; joins contain predicates; dialect-sensitive constructs are supported.
    \item \textbf{Live execution:} query planning and bounded execution succeed on the designated database.
    \item \textbf{Result quality:} output columns and row bounds match the blueprint; the relation is non-empty and non-degenerate where required.
    \item \textbf{Consistency:} instruction, program, and expected result agree; targeted mutations should not preserve the verified result.
    \item \textbf{Trajectory protocol:} tool arguments are valid, access is read-only, the budget is respected, and the final full SQL was executed before submission.
\end{enumerate}

\section{Generation Diagnostics}

\begin{table}[H]
\caption{Verified trajectories retained after three teacher rollouts per task.}
\label{tab:generation-diagnostics}
\centering
\small
\begin{tabular}{lrr}
\toprule
Source & Verified trajectories & Share \\
\midrule
Spider~2.0-Lite & 1,797 & 71.5\% \\
Spider~2.0-DBT & 715 & 28.5\% \\
\midrule
Combined & 2,512 & 100.0\% \\
\bottomrule
\end{tabular}
\end{table}

\end{document}